\documentclass{ametsocV6.1}
\usepackage{longtable}
\providecommand{\internallinenumbers}{}

\title{Can Large Language Models Review the Scientific Literature? \\
A Test Against a 102-Article Benchmark}

\authors{Abdullah Al Kalbani\aff{a,b}\correspondingauthor{Abdullah Al Kalbani, alkalbani0@outlook.com},
David M. Schultz\aff{a,c},
Alexandra K. Anderson-Frey\aff{d},
Zhonghua Zheng\aff{a,c},
and Catherine McGuire\aff{e}}

\affiliation{\aff{a}{Department of Earth and Environmental Sciences, The University of Manchester, Manchester, United Kingdom}\\
\aff{b}{Department of Computer Science, The University of Manchester, Manchester, United Kingdom}\\
\aff{c}{Centre for Crisis Studies and Mitigation, The University of Manchester, Manchester, United Kingdom}\\
\aff{d}{Department of Atmospheric and Climate Science, University of Washington, Seattle, Washington, United States}\\
\aff{e}{Research IT, The University of Manchester, Manchester, United Kingdom}}

\abstract{Large language models (LLMs) can be used to write literature reviews, but how good are they at that task?  Previous evaluations have tested steps of that task rather than the whole. This study follows the process of using LLMs to write a literature review in four steps: finding the relevant articles, interpreting each one, synthesizing across them, and proposing future work. This study uses a benchmarked literature review that was manually constructed, written, and published---a literature review on the Spanish-plume weather pattern created from a 102-article database derived from a systematic search by Schultz et al.\ (2025). In the first step, five AI tools searched for those articles. The best tool found only 42 of the 102, and together the five found 57. Of 18 articles with substantial mentions of the Spanish plume, the tools found between 4 and 12. Library access mattered: With a university library login, two coding agents found 82 and 79 articles, against 42 and 15 without. In the second step, three tools were then given PDFs of all 102 articles. All three correctly identified the 18 substantial articles, but they misclassified between 40\% and 79\% of the 84 passing mentions as substantial. In the third step, the same three tools each wrote a synthesis of the literature based on ten questions. Two expert reviewers scored the syntheses without knowing which tool had written which. They found that the syntheses were largely accurate, but found little insight from them. In the fourth step, all three tools proposed topics broader than just the Spanish plume, with two of the tools proposing future research topics close to those in the benchmark review. Our results show that no single tool was the best for all purposes.
We make eight recommendations for researchers seeking to use AI tools in the literature review process: search with subscription access and review where and why the tool was blocked; treat an AI-written synthesis as a list of claims to verify, not as a draft; have an expert review the output; check whether some publishers' journals are missing from the results; check whether older articles are missing; judge the importance of each article yourself; write a detailed prompt; and do not take a tool's confidence as a sign that it is right. Although these tools can assist a researcher in finding and reading the literature that they can access, the tools tested in this study did not synthesize the literature effectively.}

\begin{document}

% Suppress the AMS template notice that \maketitle puts in the first-page header.
\let\origlhead\lhead
\renewcommand{\lhead}[1]{\origlhead{}}
\maketitle
\let\lhead\origlhead

%\statement
%Researchers can use chatbots and coding agents to review the literature in a discipline. Whether the results of that survey are complete, accurate, and useful needs to be tested. This study tests five different AI tools on a body of literature whose full extent is known through four steps of a review rather than one: finding the articles, interpreting them, synthesizing them, and proposing future work. None of the tools found more than 41\% of the 102 articles initially, but with library access, found 80\%. When given the complete literature, the tools wrote accurate but shallow summaries. These tools can help researchers find literature, but they cannot be trusted to review it. They are research assistants, not researchers, and their outputs must be independently checked. A review is only as good as the researcher who checks it.

%%%%%%%%%%%%%%%%%%%%%%%%%%%%%%%%%%%%%%%%%%%%%%%%%%%%%%%%%%%%%%%%%%%%%
\section{Introduction}

Creating a literature review is time-consuming, and the parts that take the most time are also the ones that cannot be skipped: finding all relevant sources, reading each article carefully to obtain an accurate reflection of their results, synthesizing across the body of literature, and deciding what future research directions are worthwhile. Large language models (LLMs) now offer to do all four steps. Authors are already using them \citep[e.g.,][]{Lieberum2025}, journals are already publishing the results \citep[e.g.,][]{Schiller2024, Scherbakov2025}, and researchers have begun to describe recognizable failure modes in AI-written reviews
%researchers and journal editors have begun to describe the impacts of generative AI on the publishing process \citep[e.g.,][]{Poe2026,Thorp2026,Gartenberg2026,Siler2026}.  
\citep[e.g.,][]{Zhao2025,Poe2026}.  Thus, we are in what has been called the ``LLM-supported'' era of reviews \citep{Hauser2025}. But, how effective are these AI tools at writing literature reviews?

Previous evaluations have tested only parts of this task such as the references a tool generates, the reliability of its answers, and the fidelity of its citations \cite[e.g.,][]{Tang2025, Wu2026, Zhang2025}.  At the time of the writing of this article (October 2026), the results are mixed.  
%For retrieval
For example, in a systematic review of LLMs conducting systematic reviews in health research, authors evaluated LLM use as promising in 54\% of the 37 studies, neutral in 24\%, and nonpromising in 22\% \citep{Lieberum2025}.
More specifically, \citet{Zhang2025} used GPT-4 Turbo to answer eight specific questions that were manually checked against the actual answers. They found 86--98\% accuracy. In contrast, \citet{Wu2026} evaluated three LLMs and found accuracy rates
%(100\% minus the failure rate) 
of 20--88\%.
%for synthesizing
%For future research directions
We are unaware of studies that compare different tools on the same corpus of literature and follow the steps in the process of reviewing the literature and writing a literature review.
%(finding the articles, interpreting them one at a time, synthesizing across them and proposing what remains to be done),. 

%That is what this study does, and the four tasks organize what follows.

Moreover, evaluating whether the tools deserve our trust is harder than it looks. 
%These steps are necessary but do not address the most interesting and perhaps most important part of a literature review. 
A literature review may cite real unhallucinated articles and still misinterpret them \citep[e.g.,][]{Tang2025, Wu2026}.   Testing the ability of an AI tool to synthesize requires a corpus whose contents are known in full (so that omissions can be counted) and reviewers who know the literature well enough to spot a plausible-sounding error. Those two things rarely coincide in recent studies.  Our present study has both. 

\begin{figure}[ht]
\begin{center}
\noindent\includegraphics[width=\textwidth]{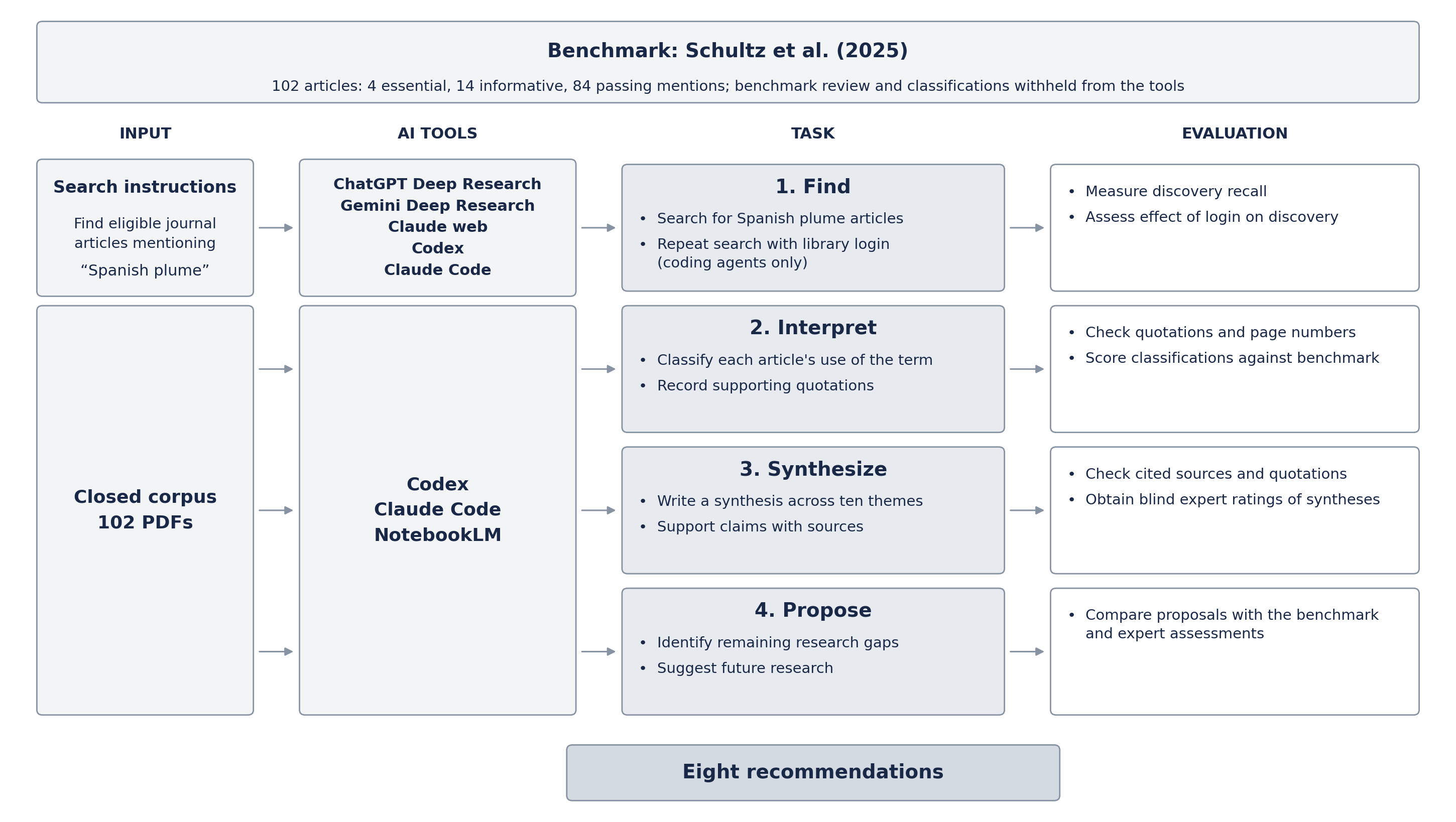}
\end{center}
\caption{Workflow of the study, showing the inputs, four tasks, and their evaluation. The closed corpus contains the same 102 source-article PDFs for interpretation, synthesis, and research-gap proposals. The benchmark review and its classifications were withheld from the tools and used for evaluation.}\label{f:flowchart}
\end{figure}

Others have found success by breaking the process of writing literature reviews down into steps that can be constructed and evaluated separately \citep[e.g.,][]{Agarwal2024} and by comparing AI tool--generated literature reviews to ones written by humans \citep[e.g.,][]{Mostafapour2024}.  We also follow that approach (Fig.~\ref{f:flowchart}).  To evaluate different LLMs across four steps in the literature-review process, we choose a baseline scientific literature review that has been human-written and already been published. \citet{Schultz2025} conducted a systematic search of the literature on the Spanish plume, a weather pattern associated with severe convective storms in western Europe. Their corpus contained 102 articles \citep{Schultz2024}, classified according to how each used the term and how central it was to the article. Then, two coauthors of the present study (DMS, the lead author of that review, and AKAF, a convective-storms expert, but not a coauthor on that review) evaluated LLM-written syntheses of the same corpus, blinded to which tool had produced which.  We then evaluate the success of six AI tools through four steps in the literature-review process using various quantitative and qualitative metrics.

This article is organized as follows.  
Section~\ref{sec:data} describes the benchmark review, the corpus, and the tools. Sections~\ref{sec:find}--\ref{sec:propose} take the four steps in turn, each with the previous work that bears on it, the method, the results and what they mean. Specifically, 
section~\ref{sec:find} asks whether an LLM can find the relevant articles. Section~\ref{sec:interpret} asks whether it can interpret each one. Section~\ref{sec:synth} asks whether it can synthesize across them. Section~\ref{sec:propose} asks whether it can propose concrete future work. 
Section~\ref{sec:recs} draws recommendations for researchers using these tools, Section~\ref{sec:limits} discusses the limitations, and Section~\ref{sec:conclusions} concludes this article.

%%%%%%%%%%%%%%%%%%%%%%%%%%%%%%%%%%%%%%%%%%%%%%%%%%%%%%%%%%%%%%%%%%%%%
\section{Data and methods}\label{sec:data}

\subsection{The benchmark review}

\citet{Schultz2025} searched Google Scholar, Web of Science, American Meteorological Society journals, and Royal Meteorological Society journals for the exact phrase ``Spanish plume'', and assembled 102 peer-reviewed articles published between 1968 and 2023. They classified each by how the term was cited. Eighty-four articles mentioned the plume in passing. Fourteen, which we call informative, used it as an integral part of their argument. Four, which we call essential, were specifically about it: \citet{Carlson1968}, who first described the mechanism; \citet{Morris1986}, who popularized the name, and both \citet{Gray1998} and \citet{Lewis2010}, who studied the climatology and synoptic weather patterns associated with Spanish plume events. Together, these 18 articles represented the substantial articles. The review also recorded, for each article, whether the term referred to the elevated dry airstream, to the broader synoptic weather pattern, or to both, and it documented the confusion that has followed from that ambiguity.

Three properties make the \citet{Schultz2025} review a good benchmark. First, the corpus is bounded, so the share of articles found is measurable. Second, the articles span several publishers with different access rules, allowing us to examine what these AI tools can access. Third, one of the central findings of the review is a terminological one: The same term ``Spanish plume" has been used for different things, which is the kind of ambiguity that requires careful understanding to catch, making it a good test of the AI tools.

\subsection{The corpus}

The 102 articles were retrieved as PDFs and assigned identifiers SP001--SP102 in order of publication year. Every file was checked for a machine-readable text layer; none required optical character recognition. The corpus and its metadata were prepared before the synthesis experiment (section~\ref{sec:synth}) and kept unchanged afterwards. The benchmark review \citep{Schultz2025}, its supplementary database \citep{Schultz2024}, and its classifications were withheld from every tool in every experiment.

\subsection{The tools}\label{sec:systems}

Six tools were used, consisting of three types: three online assistants, two coding agents, and a retrieval-grounded assistant. First, an online assistant answers in a chat interface and searches the web through the product's own tooling: ChatGPT Deep Research, Gemini Deep Research, and Claude's web interface. All three were run in their deep-research modes (Deep Research in ChatGPT and Gemini, Research in Claude), in which the assistant plans and carries out many web searches before writing a report. Second, a coding agent runs on the user's own machine with a terminal, a file system, and a browser that carries the user's sessions: Codex and Claude Code. Third, a retrieval-grounded assistant answers only from documents uploaded to it: NotebookLM, renamed Gemini Notebook in July 2026. NotebookLM was also asked to search, but that search used the benchmark's public dataset despite an instruction not to, so it was not scored (Section~\ref{sec:limits}). Table~\ref{t:systems} lists the underlying model, the plan, and which of the four tasks each tool performed. All six were used on the entry-level paid plan of their provider, and wherever a product offered a reasoning-effort setting, it was set to its maximum for every task. In Sections~\ref{sec:interpret}--\ref{sec:propose}, ``Claude'' refers to Claude Code.

\begin{table}[htbp]
\caption{The tools used, their underlying models at the time of the runs, and the tasks each performed. Model versions are those the products reported; NotebookLM does not report its model version; Gemini 3.5 is from Google's announcement \citep{Google2026}. Plan is the subscription each tool was used on. All are the provider's entry-level paid plan, which sets usage limits and, for the deep-research products, how much search each run may do. The tasks find, interpret, synthesize, and propose are represented by sections~\ref{sec:find}, \ref{sec:interpret}, \ref{sec:synth}, and \ref{sec:propose}, respectively.}\label{t:systems}
\begin{center}
\begin{tabular}{lllll}
\hline\hline
Tool & Underlying model & Type & Plan (paid) & Tasks \\
\hline
ChatGPT Deep Research & GPT-5.6 Sol & online assistant & ChatGPT Plus & find \\
Gemini Deep Research & Gemini 3.1 Pro & online assistant & Google AI Pro & find \\
Claude (web interface) & Claude Opus 5 & online assistant & Claude Pro & find \\
Codex & GPT-5.6 Sol & coding agent & ChatGPT Plus & find, interpret, synthesize, propose \\
Claude Code & Claude Opus 5 & coding agent & Claude Pro & find, interpret, synthesize, propose \\
NotebookLM & Gemini 3.5 & retrieval-grounded assistant & Google AI Pro & interpret, synthesize, propose \\
\hline
\end{tabular}
\end{center}
\end{table}

%%%%%%%%%%%%%%%%%%%%%%%%%%%%%%%%%%%%%%%%%%%%%%%%%%%%%%%%%%%%%%%%%%%%%
\section{Can LLMs find relevant articles?}\label{sec:find}

\subsection{Previous work}

\citet{Jain2024} found that five different LLMs were able to find 4–48\% of the articles from the total set of references from all five LLMs combined. \citet{Poe2026} described a ``missing middle'' in AI-written literature reviews: a handful of foundational articles from the 1980s and early 1990s, a cluster of very recent ones since 2023, and little in between (i.e., mid 1990s to early 2020s), which they attributed to publisher firewalls. That claim is testable only against a literature whose full extent is known, and Section~\ref{sec:find}\ref{sec:finddisc} tests it. The mechanism they proposed, that a model finds what is openly available and stops at the paywall, also predicts that what a tool finds should depend on the publisher and the access route rather than on the model, which Section~\ref{sec:find}\ref{sec:findresults} examines directly.

\subsection{Method}
\label{sec:findmethod}

Five tools were asked to find the peer-reviewed literature that mentions the exact phrase ``Spanish plume'', using the same prompt, without access to the benchmark review or anything derived from it: ChatGPT Deep Research, Codex, Gemini Deep Research, Claude (web interface), and Claude Code. The prompt named the four databases used by the benchmark, required each result to carry evidence that the phrase appears in the article, and prioritized finding as many articles as possible. 
%Model versions were GPT-5.6 Sol for the two OpenAI tools, Gemini 3.1 Pro, and Claude Opus 5. 
Results were matched to the corpus by DOI and normalized title. The saved search results were produced in late July and early August 2026. Institutional authentication and browser state were not controlled uniformly across these five runs, so the comparison describes what each tool reported in practice rather than a matched test of model capability. The Codex list used here is the 8 August cookie-cleared run, not its earlier search report. An article was counted as found only when the tool's report named it; a total that a tool reported without naming the articles was recorded separately.

A paired test then isolated the effect of access. Codex and Claude Code were each run twice with an identical prompt: once with browser cookies cleared and no institutional session, and once after the first author signed in to the University of Manchester's publisher and database subscriptions.

\subsection{Results}\label{sec:findresults}

All five tools found fewer than half of the 102 articles (Table~\ref{t:discovery}). Codex found 42, ChatGPT Deep Research 40, Gemini Deep Research 16, Claude Code 15 and Claude's web interface 7. Gemini reported that 35 articles passed its screening of 277 records, but its report named only 16 of them, so the other 19 could not be checked against the benchmark. Together, the five tools found 57 different articles. The other 45 were not found by any tool, although some of them may be among Gemini's 19 unnamed articles.

\begin{table}[htbp]
\caption{Articles found by each tool in searches run in late July and early August 2026. ``Reported by the tool'' is the tool's own count after screening; ``Found'' is the number of those that the tool named and that are among the 102 articles. Gemini did not name all 35 articles it reported. Browser and login state were not controlled uniformly across tools.}\label{t:discovery}
\begin{center}
\begin{tabular}{lccc}
\hline\hline
Tool & Type & Reported by the tool & Found \\
\hline
Codex (GPT-5.6 Sol) & coding agent & 42 & 42 \\
ChatGPT Deep Research (GPT-5.6 Sol) & online assistant & 40 & 40 \\
Gemini Deep Research (Gemini 3.1 Pro) & online assistant & 35 & 16 \\
Claude Code (Claude Opus 5) & coding agent & 15 & 15 \\
Claude web interface (Claude Opus 5) & online assistant & 7 & 7 \\
\hline
Found by at least one tool & & --- & 57 \\
Not found by any tool & & --- & 45 \\
\hline
\end{tabular}
\end{center}
\end{table}

\begin{figure}[htbp]
\noindent\includegraphics[width=\textwidth]{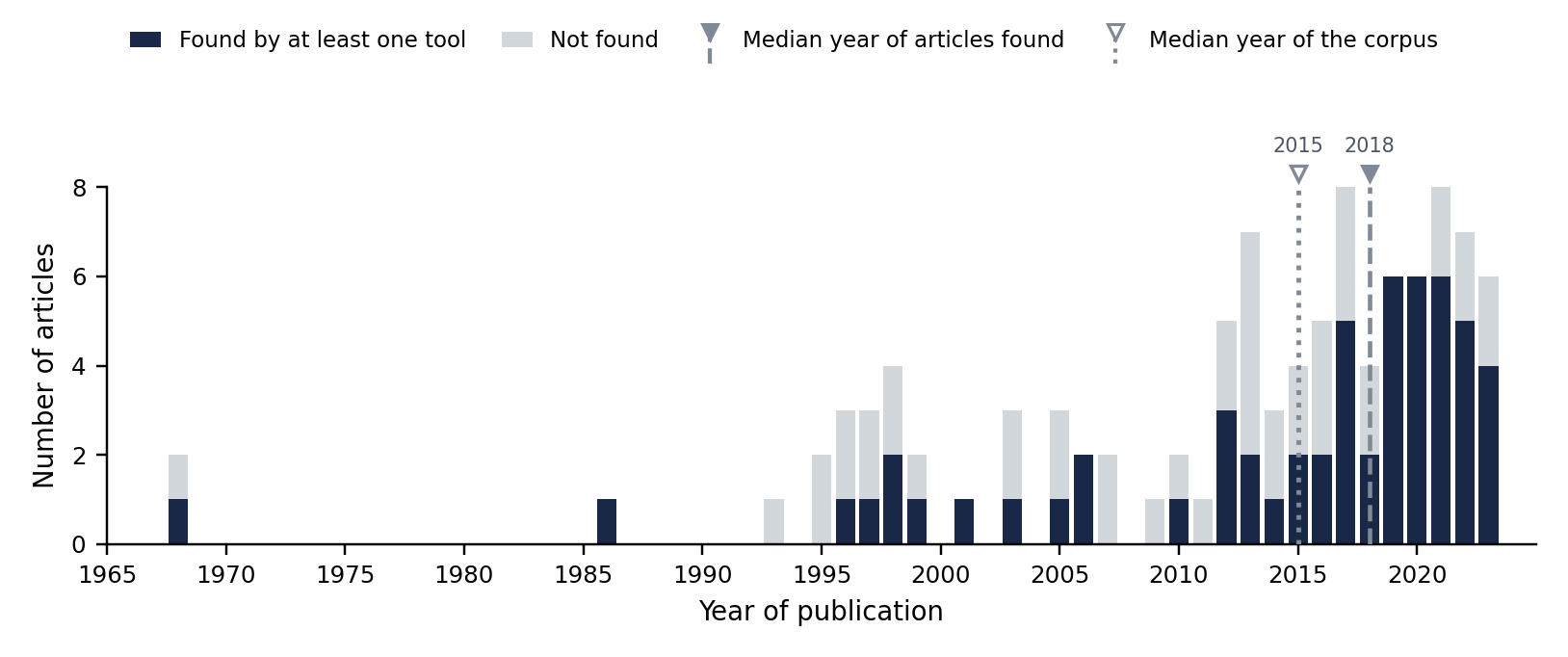}
\caption{Articles found (black) and not found (light gray) by at least one of the five tools, by publication year. The stacked bars show the 102-article corpus of \citet{Schultz2025}. The dashed line and filled triangle mark the median year of the articles found, and the dotted line and open triangle that of the corpus as a whole. Some of the articles shown as not found may be among those that Gemini reported but did not name.}\label{f:discovery}
\end{figure}

Figure~\ref{f:discovery} compares the articles found with the benchmark's year distribution. The tools found 78\% of the articles published from 2020 onward but only 48\% of those from 1995 to 2019, so their results were weighted toward recent work.

The share of articles found varied by journal (Table~\ref{t:journals}). Articles in American Meteorological Society journals are free to read one year after publication, and the tools found 10 of the 11 in the corpus, including all 5 in \textit{Weather and Forecasting} and 3 of the 4 in \textit{Monthly Weather Review}. The tools found 13 (37\%) of the articles in \textit{Weather}, which holds 35 of the 102, and 2 of the 3 in \textit{Meteorological Applications}. Only 2 of the 11 pre-2005 \textit{Weather} articles were found, against 7 of the 11 pre-2005 articles in other journals. This bears on the ``missing middle'' that \citet{Poe2026} described, which Section~\ref{sec:find}\ref{sec:finddisc} discusses.

\begin{table}[htbp]
\caption{Share of articles found, by journal and publisher, for journals holding at least three benchmark articles. ``Articles found'' counts articles that at least one tool found; articles that Gemini reported but did not name could increase these numbers. American Meteorological Society is abbreviated AMS, Royal Meteorological Society is abbreviated RMetS, and American Geophysical Union is abbreviated AGU. RMetS and AGU journals are published by Wiley.}\label{t:journals}
\begin{center}
\begin{tabular}{llrrr}
\hline\hline
Journal & Publisher & Articles in corpus & Articles found & Percentage \\
\hline
{\it Weather and Forecasting} & AMS & 5 & 5 & 100\% \\
{\it International Journal of Climatology} & Wiley (RMetS) & 6 & 6 & 100\% \\
{\it Natural Hazards and Earth System Sciences} & Copernicus & 3 & 3 & 100\% \\
{\it Monthly Weather Review} & AMS & 4 & 3 & 75\% \\
{\it Quarterly Journal of the Royal Meteorological Society} & Wiley (RMetS) & 13 & 9 & 69\% \\
{\it Meteorological Applications} & Wiley (RMetS) & 3 & 2 & 67\% \\
{\it Atmospheric Research} & Elsevier & 7 & 4 & 57\% \\
{\it Weather} & Wiley (RMetS) & 35 & 13 & 37\% \\
{\it Journal of Geophysical Research: Atmospheres} & Wiley (AGU) & 3 & 0 & 0\% \\
\hline
%\hline
%{\it Weather and Forecasting} & AMS & 5 & 5 & 100\% \\
%{\it International Journal of Climatology} & Wiley (RMetS) & 6 & 6 & 100\% \\
%{\it Natural Hazards and Earth System Sciences} & Copernicus & 3 & 3 & 100\% \\
%{\it Monthly Weather Review} & AMS & 4 & 3 & 75\% \\
%{\it Journal of Geophysical Research: Atmospheres} & Wiley (AGU) & 3 & 0 & 0\% \\
%{\it Quarterly Journal of the Royal Meteorological Society} & Wiley (RMetS) & 13 & 9 & 69\% \\
%{\it Atmospheric Research} & Elsevier & 7 & 4 & 57\% \\
%{\it Weather} & Wiley (RMetS) & 35 & 13 & 37\% \\
%{\it Meteorological Applications} & Wiley (RMetS) & 3 & 2 & 67\% \\
%\hline
\end{tabular}
\end{center}
\end{table}

The tools met different barriers along the way. Web of Science was unavailable to both coding agents without a university subscription. Google Scholar interrupted Codex with a CAPTCHA and Claude Code after 15 results; AMS blocked Claude Code entirely but admitted Codex. Access was not the only problem; search engines also differ in what they expose to search: Web of Science covers only titles, abstracts and keywords, whereas Google Scholar covers the full text. For ``Spanish plume'', Web of Science returned 11 results and Google Scholar 183.

\begin{figure}[t]
\noindent\includegraphics[width=\textwidth]{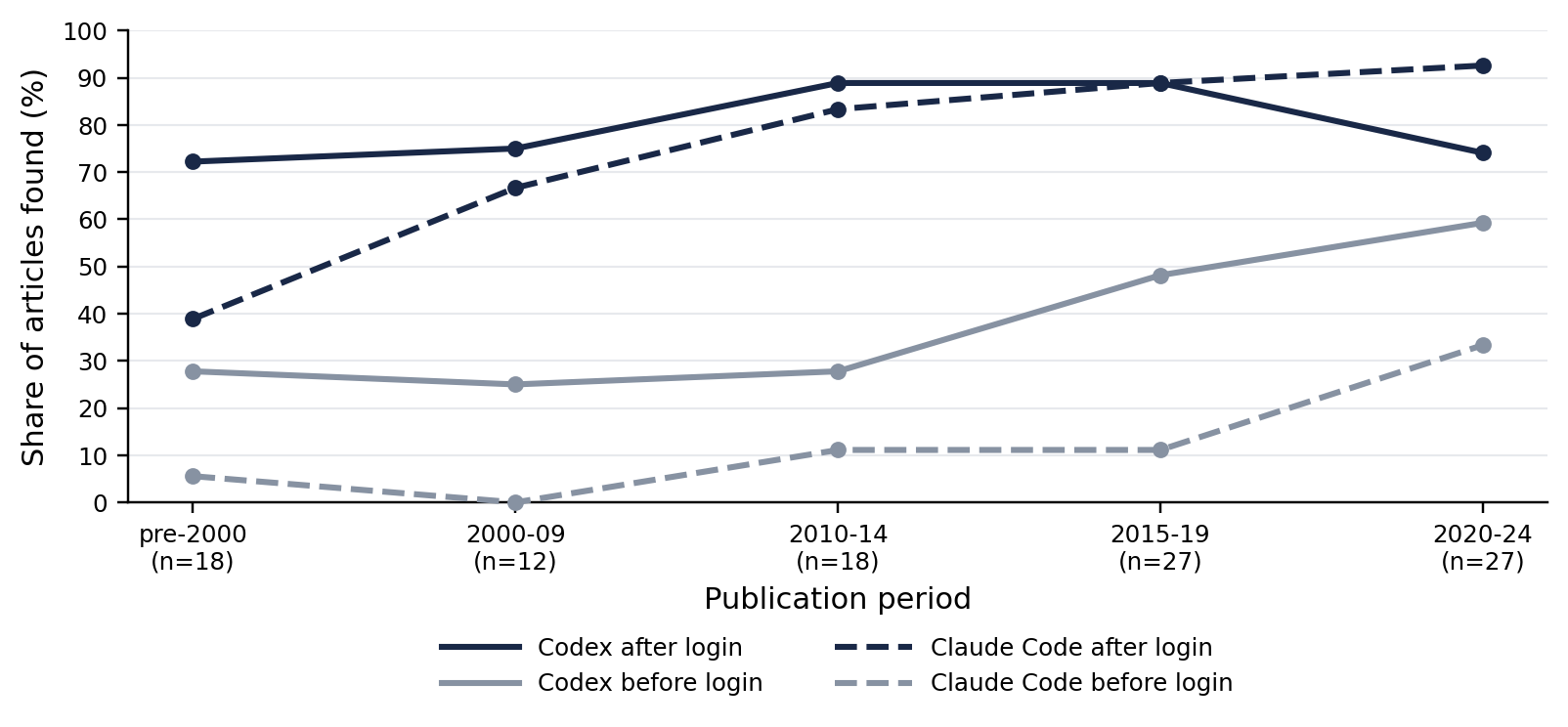}
\caption{Share of each publication period's benchmark articles found by Codex (solid) and Claude Code (dashed), before (gray) and after (black) library login. The number of articles in each period is given below its label.}\label{f:accessage}
\end{figure}

In the paired test, Codex went from 42 benchmark articles before login to 82 after, and Claude Code from 15 to 79 (Fig.~\ref{f:accessage}). Codex found none of the pre-1995 articles before login and 75\% after.

\begin{figure}[t]
\noindent\includegraphics[width=\textwidth]{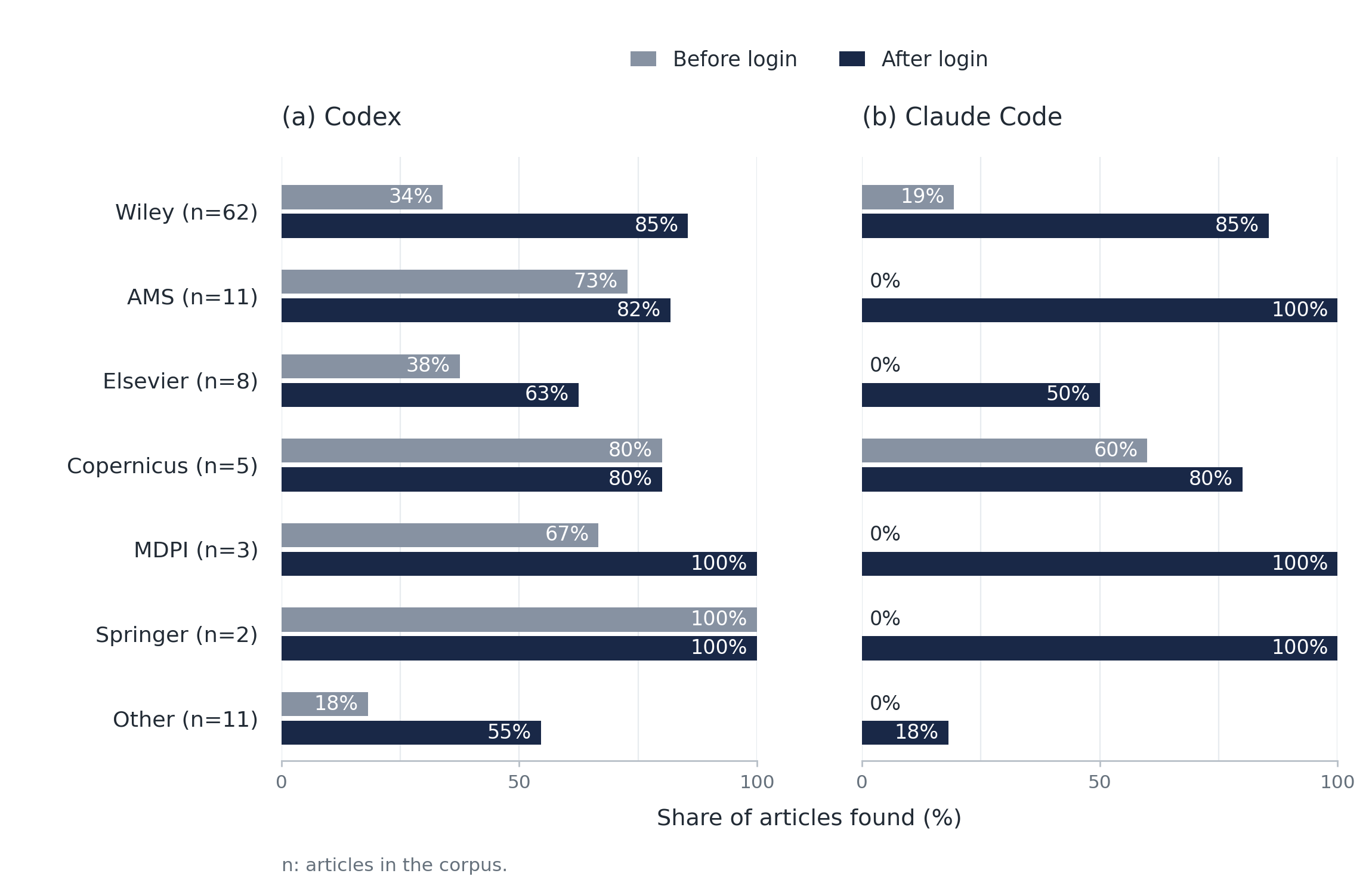}
\caption{Articles found by (a) Codex and (b) Claude Code before and after login, by publisher. Bars and labels show the percentage of each publisher's corpus articles found; n is the number of articles in that publisher group. Wiley includes the Royal Meteorological Society journals (57 articles) and five articles in other journals on its platform. 
%Groups of fewer than five articles should be read with care. 
The publisher groups do not indicate whether an article is free to read.}\label{f:publisherlogin}
\end{figure}

Without library access, Codex found 28\% of the articles published before 2000 and 59\% of those from 2020 onward, and Claude Code 6\% and 33\%. After login, Codex found 72\% and 74\%, and Claude Code 39\% and 93\% (Fig.~\ref{f:accessage}). Figure~\ref{f:publisherlogin} shows the same comparison by publisher. Most of the gain came from Wiley's journals, where Codex went from 21 of the 62 articles to 53 and Claude Code from 12 to 53. Before login, Claude Code found no articles from five of the seven publisher groups.

\subsection{Discussion: Access, not capability, limited what the tools found}\label{sec:finddisc}

The search experiment reveals constraints imposed by publishing infrastructure as well as differences among tools. Whether a tool found an article depended on two things: whether it could reach the article, and whether it could search its text. Tools of the same type found very different numbers of articles: the online assistants found 40, 16 and 7, and the coding agents 42 and 15. Because the tools differ in more than their underlying model, including how they search and which websites admit them, these differences cannot be attributed to the models alone. Articles that mention ``Spanish plume" only in their bodies cannot be reached by a search that indexes metadata alone.

The tools were held back by four barriers outside their control. First, Web of Science requires a subscription. Second, even with a subscription, Web of Science searches only an article's title, abstract and keywords, not its body text. Third, part of the literature exists online only as images. Fourth, Google Scholar and the American Meteorological Society limit automated access. The same prompt therefore met different barriers depending on which tool carried it. Articles in American Meteorological Society journals are free to read one year after publication, yet Claude Code found none of the 11 before login, because the society's website refused its automated requests. Codex, which the site admitted, found 8 (Fig.~\ref{f:publisherlogin}). Wiley, by contrast, serves \textit{Weather} articles from before about 2005 as metadata and a reference list, with the article body available only as a scanned image, so the phrase cannot be matched inside them.

Login to the university library nearly doubled the number of articles Codex found (42 to 82) and substantially raised the number Claude Code found (15 to 79). After login, Codex found a similar share of articles from every period, but Claude Code still found only 39\% of those published before 2000 (Fig.~\ref{f:accessage}). In other words, most of what the tools missed before login was a matter of access: Given the same subscriptions a researcher has, the tools were able to find far more.

\citet{Poe2026} argued that AI-written reviews cited a few decades-old foundational articles (often from the 1980s and early 1990s) and a run of very recent ones (from 2023 to 2026), skipping the literature between (from the mid-1990s to the early 2020s), because the tools could not get behind publishers' firewalls. In the present study, the tools found 21 of the 27 articles from 2020 onward (78\%), compared with 34 of the 71 from 1995--2019 (48\%). They found three of the four essential articles; the exception, \citet{Gray1998}, was in \textit{Weather}. These values are minima, because Gemini did not name all the articles it reported. The corpus contains only four articles before 1995, limiting the comparison to the results of \citet{Poe2026}.

Bot controls, subscriptions, indexing depth, and digitization all shape what an AI agent can find. The practical implication is that a coding agent running inside an authenticated institutional session can find different articles than the same agent on the open web. Future evaluations of LLM literature searches should control that environment and require a complete, auditable article list from every tool.

%%%%%%%%%%%%%%%%%%%%%%%%%%%%%%%%%%%%%%%%%%%%%%%%%%%%%%%%%%%%%%%%%%%%%
\section{Can LLMs interpret articles?}\label{sec:interpret}

\subsection{Previous work}

\citet{Wu2025} asked
%, of medical answers, 
whether the sources a model cites actually support the claim they are attached to, and automated that check. They found that 50--90\% of the responses from seven LLMs were not fully supported by the sources cited, and that even for GPT-4o with web search, about 30\% of individual statements were unsupported. \citet{vanDort2026} traced the mechanism inside a retrieval-augmented model and found that the decision to attach a citation is separable from whether the cited passage is the source of the statement, so a well-formed citation is not by itself evidence of grounding. Both of these studies bear directly on the present study, in which every quotation a tool reported was checked against its source. \citet{Zhang2025} used GPT-4 Turbo to review the microphysics literature for a numerical weather prediction model, with structured extraction as an intermediate step, which is the form this task takes here. They found that GPT-4 Turbo could answer eight questions about journal articles with 86--98\% accuracy against human-checked answers.

\subsection{Method}

\citet{Schultz2025} built their review on a database of the literature, one row per article, assembled before any of the review was written. Three tools were asked to build the same kind of database. Given the corpus and a locked 17-column schema, each produced one row per article, with no synthesis: the role of the term in the article, the sense in which it is used, the airstream properties and diagnostic variables reported, the origin and transport claims and their evidence, the plume configuration [i.e., classic or modified Spanish plume, as per \citet{Lewis2010}], capping and initiation mechanisms, hazards, whether a prefrontal trough is described, a supporting quotation with its page, and stated limitations. 
%The plume configuration follows the two synoptic patterns of \citet{Lewis2010}, as adopted by the benchmark: a classic Spanish plume, in which a north--south-oriented trough in the jet stream approaches from the eastern Atlantic, and a modified Spanish plume, in which the trough is tilted northwest--southeast, sometimes as a cutoff low, with its cyclone over the Bay of Biscay \citep[][their Table~5 and Fig.~6]{Schultz2025}. 
Codex, NotebookLM and Claude each built a database from the same locked prompt file, in a context holding only the corpus, with no web access and no sight of the benchmark. The rows were scored against the fields of the benchmark database that record the same information, with the benchmark database treated as the gold standard: Any departure counted as an error. Every quotation was then checked against the supplied PDF by an independent script, which searched the page the tool named and then the whole document under three text-extraction modes, ignoring hyphenation, line numbering, and spacing so that only a genuine departure from the source counted as a failure. Details of the quotation check are given in the public repository (see the data availability statement).

\subsection{Results}\label{sec:interpretresults}

All three tools returned a database covering all 102 articles in the required order with no blank cells. All of Codex's and Claude's quotations appeared word for word in the articles they cited. Codex gave a quotation for 100 of the 102 articles, and every one appeared word for word on the page it named. For the other two it gave none, marking them as having no mention in the main text; in one of them, the phrase appeared only in the reference list. Claude gave a quotation for 101 articles, all of them real text from the article cited, with the page number correct for 98 and wrong for 3.

NotebookLM's quotations were the least correct. Of its 101 quotations, 94 appeared word for word in the article it cited; the other 7 were taken from the article but altered its wording, dropping or changing words without marking the change. Its page numbers were less consistent: although the prompt asked for the page of the PDF, it mixed PDF pages with printed journal pages, and many were on neither. Of its 94 word-for-word quotations, 42 were on the page it gave, counting either the PDF or the printed page, and 52 were not.

\begin{figure}[t]
\noindent\includegraphics[width=\textwidth]{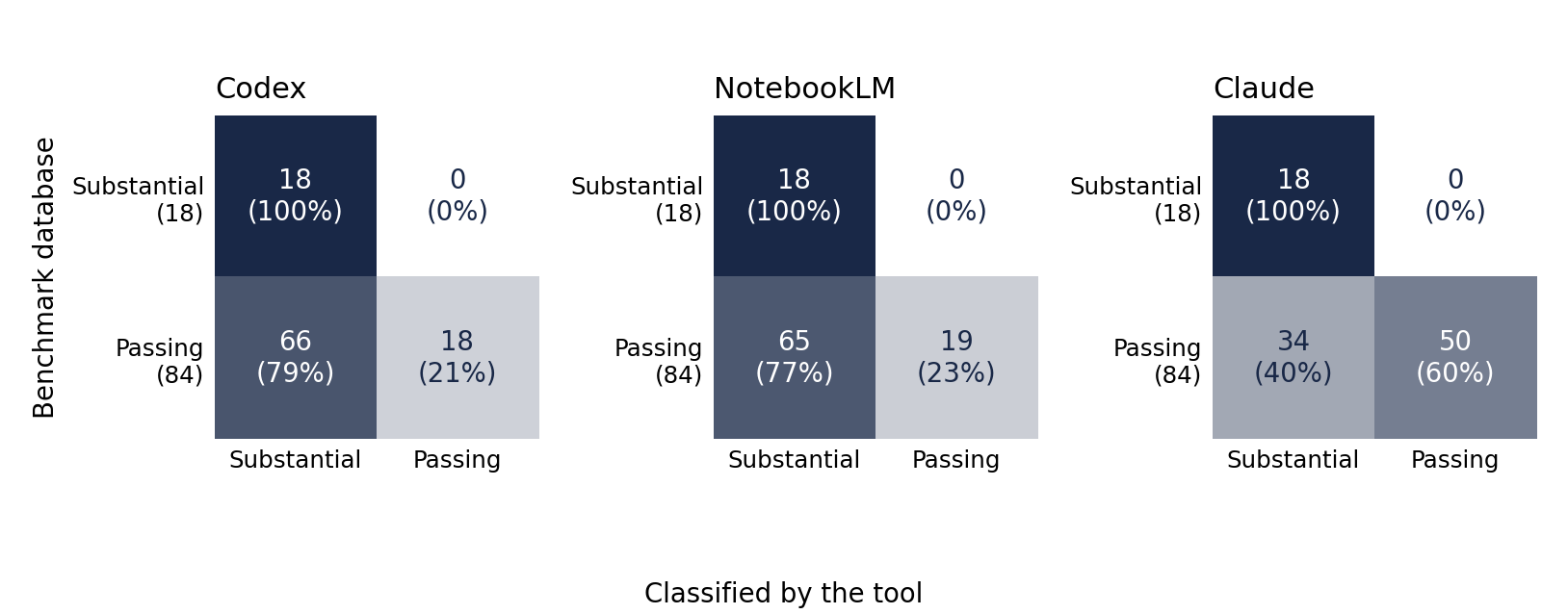}
\caption{Classification of the 102 articles as substantial or passing mention by (left) Codex, (middle) NotebookLM and (right) Claude. Rows give the benchmark database's classification and columns the tool's. Each cell gives the number of articles and, in parentheses, the share of its row; darker cells hold a larger share.}\label{f:confusion}
\end{figure}

Scored against the benchmark database, all three tools classified the 18 substantial articles correctly and misclassified many of the 84 passing mentions (Table~\ref{t:extraction} and Fig.~\ref{f:confusion}). 
%The benchmark database classified 18 articles as substantive (the four essential and fourteen informative) and 84 as passing mentions. 
%All three tools classified all 18 substantive articles correctly. 
The errors were on the passing mentions: Codex misclassified 79\% of them as substantial, NotebookLM 77\% and Claude 40\%. Cohen's $\kappa$, which measures agreement with the benchmark beyond what chance would produce, was 0.09 for both Codex and NotebookLM (i.e., little better than chance) and 0.34 for Claude (i.e., fair agreement). All three tools had a recall of 1.00 for the substantial articles but a precision of only 0.21 to 0.35, because they classed so many passing mentions as substantial. Recognizing an article about the Spanish plume was therefore easy for all three tools. Recognizing one that mentions it only in passing was harder: Claude was appreciably better at it than the other two, but it still misclassified 40\% of the passing mentions.

\begin{table}[t]
\caption{The databases built by the three tools, scored against the benchmark database as the gold standard, and the quotation audit. Accuracy on the two usage flags is the share of the 102 articles on which each flag (the term used for the airstream, for the synoptic pattern, or both) matches the benchmark database. Precision is the share of the articles a tool classed as substantial that the benchmark database also classes as substantial, recall is the share of the 18 substantial articles that the tool classed as substantial, and F1 is the harmonic mean of the two. Configuration (classic or modified plume) is scored only on the eight articles for which the benchmark database gives one; for one of them it gives both, so either answer counts as correct. The benchmark database marks a prefrontal trough in seven articles. The quotation rows are the first author's own check against the supplied PDFs, not the tools' self-reports. How the quotations were matched is described in the public repository.}\label{t:extraction}
\begin{center}
\begin{tabular}{lccc}
\hline\hline
 & Codex & NotebookLM & Claude \\
\hline
Substantial articles correct (of 18) & 18 & 18 & 18 \\
Passing mentions misclassified (of 84) & 66 & 65 & 34 \\
Passing mentions correct (of 84) & 18 & 19 & 50 \\
Accuracy on passing/substantial & 35\% & 36\% & 67\% \\
Cohen's $\kappa$ on that split & 0.09 & 0.09 & 0.34 \\
Precision (substantial class) & 0.21 & 0.22 & 0.35 \\
Recall (substantial class) & 1.00 & 1.00 & 1.00 \\
F1 (substantial class) & 0.35 & 0.36 & 0.51 \\
Airstream sense: accuracy & 65\% & 71\% & 70\% \\
Synoptic-pattern sense: accuracy & 71\% & 64\% & 75\% \\
Configuration correct (of 8) & 6 & 7 & 7 \\
Prefrontal trough: correct (of 7) & 1 & 6 & 4 \\
Prefrontal trough: false positives & 0 & 33 & 10 \\
\hline
Quotations offered (of 102) & 100 & 101 & 101 \\
Word for word & 100 & 94 & 101 \\
Wording altered & 0 & 7 & 0 \\
\hline
\end{tabular}
\end{center}
\end{table}

The tools also recorded whether an article used ``Spanish plume'' to mean the airstream, the synoptic weather pattern, or both, and they agreed with the benchmark database for about two-thirds to three-quarters of the articles (Table~\ref{t:extraction}). The disagreements went both ways: Codex and Claude assigned the airstream meaning to many more articles than the benchmark database did, and NotebookLM to many fewer. Where the benchmark database recorded a classic or a modified Spanish plume, the tools mostly agreed with it.
%For the situation where the Spanish plume was incorrectly referred to as a prefrontal trough \citep[][p. 742]{Schultz2025}, the more articles in which a tool identified prefrontal troughs in the benchmarked articles, the more articles is mistakenly attributed the Spanish plume to being a prefrontal trough.
%there was a trade-off: 
%the more articles a tool marked as describing a prefrontal trough, the more of the benchmark's cases it found and the more articles it marked wrongly.

\subsection{Discussion: Accurate extraction, unreliable interpretation}

The tools extracted text from the articles accurately, but they were unreliable at judging what an article did with the term. All three tools recognized the 18 articles that discuss the Spanish plume substantially, yet they also classified between 40\% and 79\% of the passing mentions as substantial, implying the tools often mistook a mention of the term for a discussion of it. For a literature review, making this distinction matters, because the tools would overstate the extent to which the literature discusses the topic in depth.

Quotation accuracy depended on the tool. Codex and Claude quoted word for word in every row where they gave a quotation, whereas 7 of NotebookLM's 101 quotations altered the article's wording. NotebookLM was also the tool whose two quotations in its synthesis were misattributed (Section~\ref{sec:synth}\ref{sec:synthresults}). The database made these checks easy. Each row gave a quotation with its article and page, so a script could check all of them. In a written synthesis, the quotations were spread throughout the text, and each one had to be found and checked by hand. A researcher who asks a tool for a synthesis may therefore want a database from it, as well, because the database can be checked automatically.

%%%%%%%%%%%%%%%%%%%%%%%%%%%%%%%%%%%%%%%%%%%%%%%%%%%%%%%%%%%%%%%%%%%%%
\section{Can LLMs synthesize across articles?}\label{sec:synth}

\subsection{Previous work}

\citet{Tang2025} evaluated LLMs on reference generation, abstract writing and review composition. They found that at most 52\% of the references the LLMs generated matched a real article. \citet{Wu2026} proposed a benchmark for the reliability of LLMs in searching for and reading papers. They found that three LLMs missed more than 60\% of the relevant literature when asked to find articles on a topic. \citet{Abdulhai2026} found that when AI tools were asked only to correct grammar, they shifted the meaning of the text they edited, which is the failure a synthesis risks at corpus scale. These three studies measured whether a model's output was true and whether its citations were real. None of them asked domain experts to judge a cross-article synthesis against a complete manual review of the same literature.

\subsection{Method}\label{sec:synthmethod}

NotebookLM, Codex, and Claude Code received all 102 PDFs and the same locked prompt. They were asked to write a critical synthesis, not an article-by-article summary, organized around ten themes:

\begin{enumerate}
\itemsep0pt
\item evolution of the term and concept;
\item airstream versus synoptic-pattern usage;
\item thermodynamic and moisture properties;
\item claimed geographical origin and transport pathways;
\item evidence used to support origin claims;
\item capping, instability, and convection-initiation mechanisms;
\item hazards, locations, and event types;
\item areas of agreement;
\item disagreements, unsupported assumptions, and uncertainty;
\item remaining research gaps.
\end{enumerate}

The prompt required a source and a page for every claim, forbade outside knowledge and web access, and instructed the tool to write ``Not established by the provided sources'' where evidence was absent. The prompt also asked each tool to label statements as a direct finding, an author interpretation, or a cross-article inference. Four of the ten themes encoded concerns raised by the benchmark review itself, so this was a guided synthesis test rather than an open-ended one. The tools were not asked to discover the problems in the literature, only to discuss them.

The prompt set the same rules for all three tools:
\begin{enumerate}
\itemsep0pt
\item Cite the supporting source and its one-based PDF page as \texttt{[SPxxx, PDF p.\ N]}.
\item Label each statement as a direct finding, an author interpretation, or a cross-article synthesis inference.
\item Weight analytical studies more heavily than passing mentions, and do not treat an appearance in a reference list as evidence that an article discusses the plume.
\item Do not convert repeated unsupported claims into established facts, and do not describe anything as consensus unless the supplied sources demonstrate it.
\item Never guess a page number, and where the evidence is insufficient, write ``Not established by the provided sources.''
\item Close with a claim ledger giving, for every theme, the claim, its supporting sources, any contradicting or qualifying sources, the strength of the evidence, and the residual uncertainty.
\end{enumerate}

The three syntheses were anonymized as Syntheses 1, 2 and 3, with the key held by the first author and withheld from both reviewers (1 was NotebookLM, 2 was Codex, and 3 was Claude). Citation formatting was standardized across the three so that style could not identify a tool; claims, hedges, identifiers and page numbers were left untouched. Two coauthors (DMS and AKAF) scored each synthesis independently, theme by theme, on five criteria developed jointly: writing quality and readability, depth and breadth, veracity, insight, and usefulness to a researcher conducting a review. Each criterion was rated from 1 (strongly disagree) to 5 (strongly agree), giving 50 ratings per synthesis per reviewer and 300 in total. Reviewer A is the lead author of the benchmark review (DMS); Reviewer B is an independent researcher in severe convection who was not involved in that study (AKAF). Both also returned written comments. Reviewer A additionally checked every direct quotation in Syntheses 1 and 2, and those in the first two themes of Synthesis 3, against the source PDFs.

\subsection{Results}\label{sec:synthresults}

The three syntheses differed in size. NotebookLM wrote 3624 words across the ten themes and cited 41 distinct corpus articles; Codex wrote 3919 words and cited 27; Claude wrote 7480 words and cited 79. None of these numbers predicted the expert ratings.

\begin{table}[t]
\caption{Mean blinded rating (1--5) per criterion, averaged over the ten themes, for each reviewer and tool. Reviewer A is the lead author of the benchmark review; Reviewer B is an independent severe-convection researcher.}\label{t:scores}
\begin{center}
\begin{tabular}{llccccc|c}
\hline\hline
Reviewer & Tool & Quality & Depth & Veracity & Insight & Utility & Mean \\
\hline
Reviewer A & NotebookLM & 4.2 & 3.3 & 4.2 & 3.4 & 3.3 & 3.68 \\
 & Codex & 3.6 & 3.0 & 3.7 & 3.2 & 3.2 & 3.34 \\
 & Claude & 4.4 & 4.3 & 4.3 & 4.2 & 4.4 & 4.32 \\
\hline
Reviewer B & NotebookLM & 2.9 & 2.6 & 3.7 & 2.3 & 2.4 & 2.78 \\
 & Codex & 3.1 & 2.9 & 3.8 & 2.4 & 2.4 & 2.92 \\
 & Claude & 2.8 & 2.9 & 3.6 & 2.5 & 2.3 & 2.82 \\
\hline
\end{tabular}
\end{center}
\end{table}

Table~\ref{t:scores} gives the mean rating per criterion. Veracity was rated higher than insight in all six combinations of reviewer and tool, but it was not always the highest-ranked criterion, and insight was not always the lowest. Reviewer B scored veracity at 3.6 to 3.8 and insight at 2.3 to 2.5 for every tool, and wrote that the outputs were ``fine, but quite uninsightful'' and ``just a summary without any insights.'' Reviewer A described the same tendency: ``Like undergraduates sometimes do, the LLM offers text that is not answering the original question.'' The tools reported the literature; they did not synthesize it.

The reviewers also converged on a failure mode. Theme 8 asked for areas of agreement in the literature, and all three tools answered with more confidence than the corpus supports. Synthesis 1 asserted ``absolute consensus'' and called the plume ``the primary driver of Europe's most severe summertime convective hazards''; Reviewer B challenged both phrases, and Reviewer A noted that all three tools ``uncharacteristically use strong language and express confidence'' on that theme, and asked whether a model recognizes degrees of certainty at all. The prompt had explicitly forbidden describing anything as consensus unless the sources demonstrated it.

Reviewer A's quotation check produced the sharpest single result. Codex used no direct quotations. NotebookLM used two, and both were wrong.
One quotation was attributed to \citet{Fery2024}, a sentence that does not appear in the article but appears word for word in a referee's comment on their manuscript, at a page number the article does not have. 
%The sentence attributed to \citet{Fery2024} , not in the article itself. 
The other quotation was from \citet{Schultz2005}, with the wrong page number. Although it was outside the quotation marks, NotebookLM changed the word ``mechanism'' in the original to ``mechanics'' in its synthesis, making the sentence nonsensical. 
%Checked afterwards against the source documents, neither quotation was invented. 
%
%The \citet{Schultz2005} words are verbatim, with ``mechanics'' outside the quotation marks, but the page is wrong. 

Claude quoted extensively, and every quotation checked was accurate, although two multi-author articles were cited by their first two authors rather than as ``et al." (e.g., ``De Meutter and Gerard'' rather than ``De Meutter et al.''). The databases gave the same picture: NotebookLM was the only tool that misquoted, and it did so in both tasks (Section~\ref{sec:interpret}\ref{sec:interpretresults}). The reviewers also found errors that no citation audit would catch: One synthesis abbreviated ``capping inversion'' as CIN, which denotes ``convective inhibition" instead; one described a warm conveyor belt bringing cooler air instead of warmer air; and one described the plume as a warm, moist layer. Several articles in the corpus describe it that way, but the description is inconsistent with the original definition of \citet{Carlson1968}, in which the plume is warm, dry air above a moist layer. The synthesis therefore reproduced an error that is present in the literature. Two tools invented or altered terminology for ``underrunning" (i.e., a process by which warm moist air flows under the capping inversion and yields the potential for convective storms to form), renaming it ``undercutting'' and coining the phrase ``edge-of-lid escape''. Reviewer A observed that none of the three recognized that Spain lies on the trajectory path from Africa, a point central to the debate about the origin of the Spanish plume airstream.

NotebookLM produced ASCII art to help answer some of these questions of varying quality and utility. In one case, the figure could have been more effective as a table.  In the second case, NotebookLM tried to produce an atmospheric sounding profile of the atmospheric layers commonly associated with the Spanish plume. This image did not translate well through ASCII art.  The third case was unintelligible.  It was not possible to understand what was being graphed nor what it was trying to convey.  No explanation of the figure or figure caption was included.

NotebookLM and Claude generally produced readable text. In contrast, the writing with Codex was the most stilted and awkward of all three LLMs (e.g., unusual sentence structure, unusual word choice), often to the point of being a negative and weakening its authority even when it did say something correct or clever.  Codex created enumerated or bulleted lists more commonly than the other two.  These lists were not necessarily things that needed ordered by a list, however.

\subsection{What the syntheses cited}\label{sec:coverage}

All three tools cited all four essential articles. Of the fourteen informative articles, Claude cited 14, NotebookLM 12, and Codex 9 (Table~\ref{t:hierarchy}). 
%Claude was the only tool to cite all 18 key articles; NotebookLM cited 16 and Codex 13. 
As a share of each tool's own citations, the eighteen substantial articles were 48\% of Codex's, 39\% of NotebookLM's and 23\% of Claude's. Citing more articles did not mean citing the right ones, however.  Even though Claude cited all 18 substantial articles, it did so by citing 77\% (79 out of 102) of the total corpus.  In contrast, Codex cited the fewest, 26\% (27 out of 102), but had the largest share of its citations on the 13 substantial articles it did cite. 

\begin{table}[t]
\caption{Corpus articles cited in each synthesis, by the benchmark's classification of how the article cites the Spanish plume.}\label{t:hierarchy}
\begin{center}
\begin{tabular}{lcccc}
\hline\hline
Tool & Essential (4) & Informative (14) & Passing (84) & Total cited \\
\hline
NotebookLM & 4 & 12 & 25 & 41 \\
Codex & 4 & 9 & 14 & 27 \\
Claude & 4 & 14 & 61 & 79 \\
\hline
\end{tabular}
\end{center}
\end{table}

\begin{figure}[t]
\centering
\includegraphics[width=\textwidth]{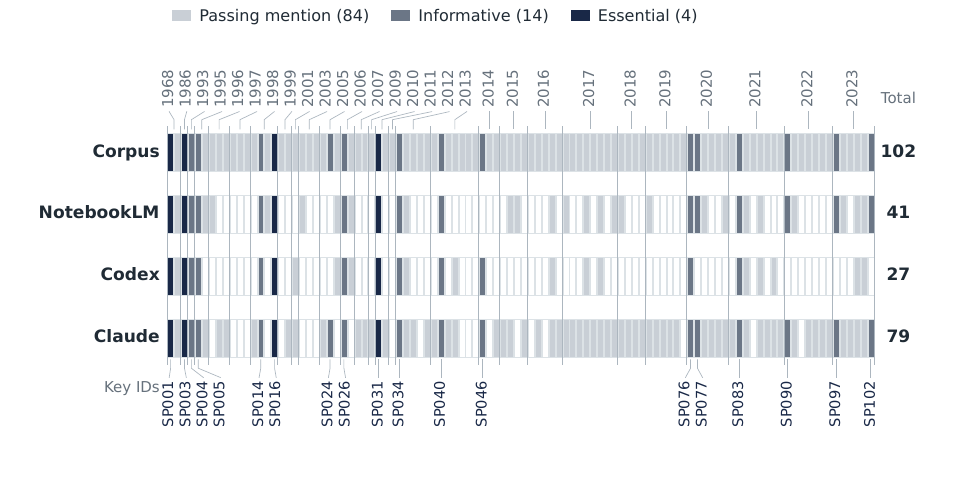}
\caption{Corpus articles cited in each synthesis. Each column is one of the 102 articles, ordered by publication year and then identifier (SP001--SP102). Colors show the benchmark classification \citep{Schultz2025}; in the tool rows, white cells are articles not cited. The totals on the right count the articles in each row. Year groups are as wide as the number of articles in them, not the time they span. Identifiers are shown only for the 4 essential and 14 informative articles.}
\label{f:keypapers}
\end{figure}

\begin{figure}[htpb]
\noindent\includegraphics[width=\textwidth]{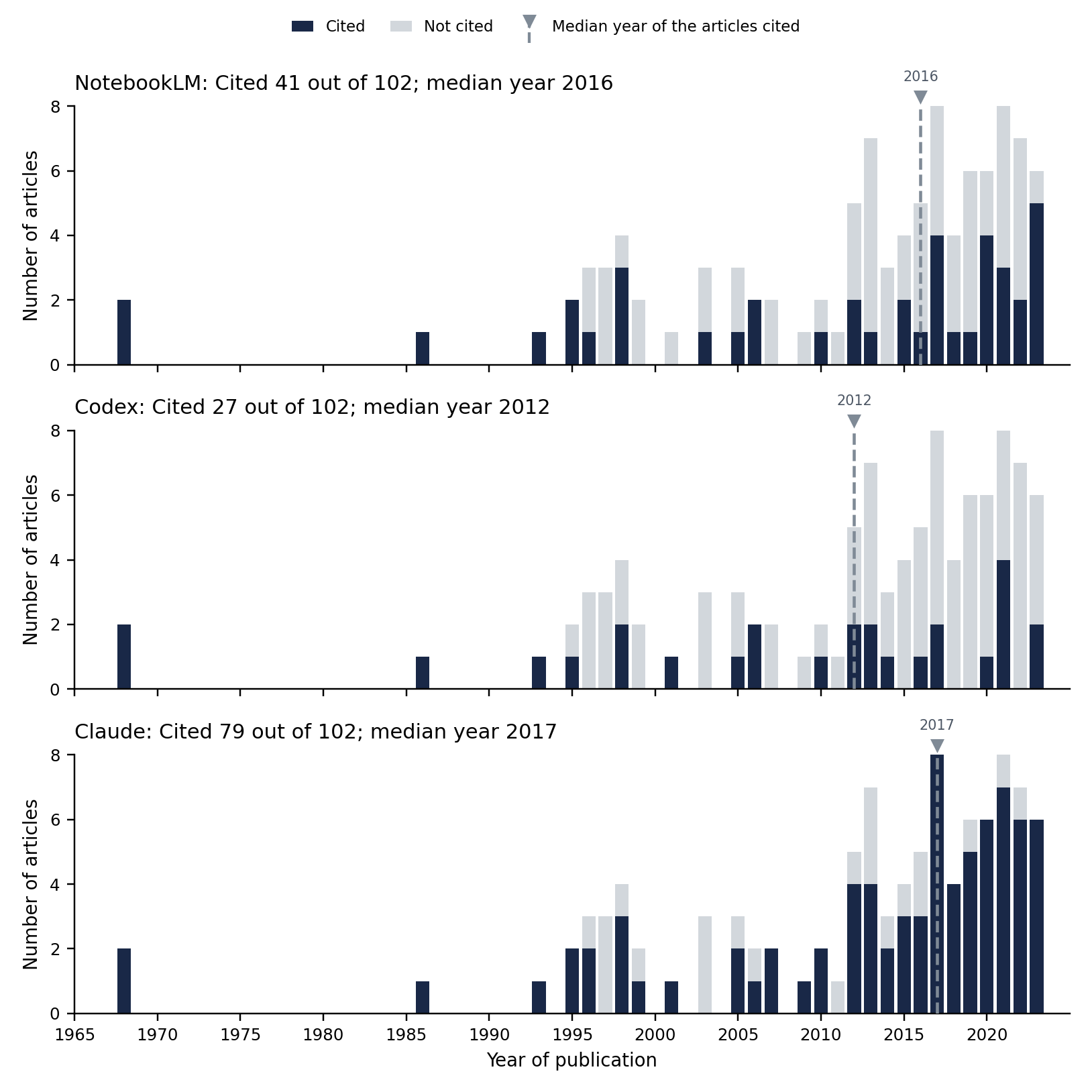}
\caption{Corpus articles cited (black) and not cited (light gray) by each synthesis, by year of publication. The dashed line and triangle mark the median year of the articles each tool cited.}\label{f:citeyears}
\end{figure}

Codex's five omissions among the informative articles were published in 2005, 2020, 2022, 2023 and 2023. It cited seven of the eight informative articles published before 2020 and two of the six published from 2020 onward (Fig.~\ref{f:keypapers}). Figure~\ref{f:citeyears} shows the three citation distributions by year of publication. Codex's lean toward older articles held across all its citations: the median year of the articles it cited was 2012, against 2016 for NotebookLM, 2017 for Claude and 2015 for the corpus. NotebookLM cited the oldest and newest articles more often than those in between, and Claude cited a larger share of the articles the more recent they were.

\subsection{Discussion: Accurate but not insightful, partly by design}

Both reviewers identified the same concerns about the tools. The syntheses got things right and did little with them. Some of that was due to the tools themselves. Some of it was due to the prompt, for which small changes have been known to produce dramatically different results \citep{Felizardo2025}. Every grounding rule in the prompt penalized an unsupported statement; none rewarded a justified interpretation. The prompt regulated quotation heavily (exact wording, verified pages, no repair), but never asked for quotation directly.  Thus, quoting became the safest way to comply, whereas paraphrasing became a veracity risk. Only one of the three labels, ``cross-article synthesis inference'', allowed a tool to reason across articles, and the prompt did not define it. Each coding agent wrote its own definition into a rules file; NotebookLM had none. Both reviewers independently found the labels unclear. Because the prompt was designed mainly to prevent errors, the tools played safe: Their syntheses were accurate and offered little insight, as the reviewers' scores show.

Reviewer B wrote the following about one of Codex's answers: ``A lot of plausible-sounding discussion that kind of circles around a whole lot of nothing." A further point was raised by Reviewer A who wrote, ``When an LLM hits upon clever insights, but then says something stupid in the next sentence or paragraph, it makes you wonder about the randomness of it all and reminds you that they are ultimately just glorified autocomplete tools.  There is no guarantee that they are on a ``hot streak” or have ``developed insight into the problem”. ...sometimes you can’t tell whether the insight is clever or not.  As per Spinal Tap: `it’s a thin line between clever and stupid.'"

\subsection{Discussion: Breadth is not quality}

Claude cited 79 articles, NotebookLM 41, and Codex 27. Reviewer A ranked Claude first, NotebookLM second, and Codex third, whereas Reviewer B ranked Codex first. This was not the order of their length: Codex wrote more words than NotebookLM. Codex covered 13 of the 18 substantial articles with 27 citations; Claude covered all 18 with 79. If completeness against the essential literature were the goal, Claude achieved it, but largely by citing most of the corpus, and Reviewer A noted that its length meant ``more opportunities to get it wrong.'' A review that cites everything has not made the judgment a review is for.

\citet{Wu2026} reported fabricated responses from an older version of Gemini. Our NotebookLM results also revealed problems with the accuracy of the quotation and the attribution of the source. Similarly, \citet{Abdulhai2026} found that LLMs changed the meaning of a text even when asked only to correct its grammar, and NotebookLM's seven altered quotations are an example. We found no invented references in the five scored searches or in the syntheses: Every article that a tool named in them exists. The excluded NotebookLM search did report one article that does not exist, with a DOI that belongs to an unrelated paper. We did find incorrect citation details, however, including wrong titles attached to real articles in two of the search reports, wrong page numbers, and a quotation attributed to an article when it came from a referee's comment.

%%%%%%%%%%%%%%%%%%%%%%%%%%%%%%%%%%%%%%%%%%%%%%%%%%%%%%%%%%%%%%%%%%%%%
\section{Can LLMs propose future work?}\label{sec:propose}

\subsection{Previous work}

Of the four tasks, this is the least studied and the hardest to score. A proposed research gap is not right or wrong in the way a citation is, so it can only be judged by researchers who are experts in the field. \cite{Zhao2025} conducted an evaluation that scored the research directions proposed by GPT-4 against the judgment of humans. They found that applying a structured series of prompts produced a literature review comparable to that written by humans and one that could identify research gaps.

\subsection{Method}

The tenth theme of the synthesis prompt (Section~\ref{sec:synth}\ref{sec:synthmethod}) asked each tool to identify the remaining research gaps in the literature, which is the fourth of the project's questions. The answers provided by the AI tools were anonymized with the rest of the synthesis and scored blind by both reviewers on the same five criteria, giving ten ratings per tool for this theme.

\subsection{Results}

NotebookLM listed four gaps and presented each as one that the articles themselves had named: that convection starting overnight above the surface is poorly understood, that prefrontal troughs have not been verified with observations, that small-scale cloud processes need better modelling, and that the effect of climate change on Spanish plume events is uncertain.

Codex presented its gaps as its own inferences from comparing articles. It proposed defining the Spanish plume, establishing the origin of the air with trajectories across many events, studying how the airstream is heated and modified over the Iberian plateau, testing how convection is initiated, studying plume events that produced no storms, building comparable records of the hazards, studying the link to meteotsunamis, and attributing changes to climate change.

Claude characterized its gaps by where they came from. It listed gaps that authors in the corpus had stated, including the origin of convergence lines and the overnight initiation of convection. It then listed gaps it inferred from what the corpus did not contain: a test for deciding whether an event is a Spanish plume, the origin of the air across many events, the inconsistency between articles that call the dry air aloft the plume and those that call the moist air beneath it the plume, a comparison of the proposed initiation mechanisms, trends in plume events, and the verification of forecasts of the Spanish plume.

Reviewer A commented that all three tools ``pull in gaps in research topics that are larger than the Spanish plume alone'', such as microphysical modelling and coastal coupling, that Claude's answer stayed the most focused, and that it was the only one to say where its gaps came from. He rated Claude's answer highest and NotebookLM's lowest. Reviewer B rated the three answers about equally, noting that NotebookLM was ``fine, but quite uninsightful."

\subsection{Discussion: Relevant gaps, but not confined to the topic}

The three answers differ in the way that they present and interpret the research needs. NotebookLM only repeated gaps that authors had already stated, whereas Codex and Claude also worked out gaps from what the literature lacked, which is what proposing future work requires.

The gaps that Codex and Claude proposed are close to the benchmark review's own recommendations. The review recommends keeping to the original definition of the Spanish plume by \cite{Carlson1968} because later articles applied the name incorrectly to a humid near-surface airstream, recommends calculating trajectories explicitly, suggests studying plume events that do not produce storms, and identifies the role of the Iberian terrain and the initiation of convection as its two research opportunities \citep{Schultz2025}. Codex and Claude reached versions of each of these. NotebookLM reached one (i.e., the initiation of convection) and only for the overnight convection.

The weakness was focus. All three tools proposed gaps on subjects broader than the Spanish plume, as Reviewer A noted. Proposing future work also depends on the step before it: A tool can only identify what is missing from a literature if it has understood what the literature contains. These results rest on one answer from each tool, scored by two reviewers.

%%%%%%%%%%%%%%%%%%%%%%%%%%%%%%%%%%%%%%%%%%%%%%%%%%%%%%%%%%%%%%%%%%%%%
\section{Recommendations}\label{sec:recs}

This study compared the results of various AI tools against a manually constructed benchmark literature review.  That comparison showed weaknesses that a researcher using these tools could not see from the tools' output alone. Someone using LLMs to construct a literature review should therefore be aware of the following issues and consider the following recommendations.
\begin{enumerate}
\item Search with subscription access to improve the results. Review the search results to see where and why the tool was blocked, by auditing its search record.
\item The tools reached some publishers' articles far more readily than others'. A gap in a tool's results may therefore reflect lack of access rather than the lack of literature. Thus, searches should also be conducted on publishers' websites and literature databases directly.
\item Recent articles were found by the tools %far 
more often than older ones. Older work may therefore be missing even when it exists, so search for it separately.
\item Judge the importance of each article yourself. Although these tools identified most, if not all, of the most important articles on the Spanish plume, there is no guarantee that such thoroughness would be expected on another topic.  Moreover, the high number of articles reported to be important by the tools, but had mere passing citations to the term ``Spanish plume", indicates that a tool cannot reliably characterize an article's importance.
\item The prompt shapes the answer, so more detailed information about the inclusion/exclusion rules for finding or more detailed information about the character of the answers expected during interpretation generally makes for better results.
\item Check the claims and quotations generated by a tool's synthesis against the original sources. As a common disclaimer on generative AI tools reads, ``AI-generated content may be incorrect".
\item A tool will not point out its own weak spots. It tends to sound confident even when it is wrong.
\item Expert review by humans is still needed. Some of a tool's errors can only be caught by someone who knows the subject.
\end{enumerate}

None of this means the tools are not useful. They can find many of the relevant articles and summarize them reasonably accurately, which could save time. But their searches have gaps, their judgments are unreliable, and their confidence is not a guide to accuracy, so their work has to be checked before it is used.

\section{Limitations}\label{sec:limits}

This study tested one topic and one set of 102 articles, and each tool was run once on each task. Differences between tools could therefore reflect chance variation between runs, which was not measured. New versions of these AI tools are being released regularly and in short order. Our results describe the output from versions of these six AI tools available in July and August 2026. Those researchers looking to see if the results change over time with future versions of these same tools can find the prompts and data in the public repository \citep{AlKalbani2026}. Although the tools were instructed not to use the benchmark review, and the results suggested they did not, we cannot know whether the underlying LLMs had been trained on it, as the review was published more than a year before our tests were performed.

The search results have several limitations. The five tools were not run under identical conditions, because login and browser state were not controlled in the same way for each. In the login test, logging in gave the two coding agents subscription access and also let them past websites that had blocked them, so the test cannot separate those two effects. A sixth search, with NotebookLM, was excluded because its sources included the benchmark's public dataset despite an instruction to exclude it. The benchmark has only four articles from before 1995, which limits what can be said about older literature.

The databases built by the tools were scored on 4 of their 17 fields, the ones that the benchmark database also records. Quotations and their page numbers were checked against the original articles, but the remaining fields, such as the origin of the air and the hazards could not be checked. The benchmark database was treated as correct, so any error in it was counted as an error by the tool. In two fields, Claude added a short justification after its answer, which the other tools did not; only the answer itself was scored.

The synthesis prompt told the tools what to write about, rather than leaving them to write openly, so we did not test whether they would have found the main problems in the literature on their own. Proposing future work was also only one of the prompt's themes, so each tool answered it once.

The results also depend on the prompts. Each task used a single prompt, and no other wording was tested. The synthesis prompt was written to prevent errors and did not ask for interpretation, it listed three labels without defining them, and it set no limit on length. The low ratings for insight, and the differences in length between the syntheses, may therefore reflect the prompt as much as the tools. Different prompts could yield different results.

The two reviewers differed. Across the 150 ratings each gave, they agreed exactly on 24\% and within one point on 67\%, Reviewer A scored 0.94 points higher on average than Reviewer B, and they ranked the three tools differently. Two reviewers are too few to assess statistical significance, and the results would be subjective anyway.  Reviewer A led the writing of the benchmark review and was potentially biased by being too close to the subject matter. The ratings are therefore given separately for each reviewer, and the conclusions rest on what both reviewers found, not on any ranking of the tools. This result is perhaps not surprising and is comparable to results from other studies that reviewers often do not agree with each other \citep[e.g.,][]{Ernst93,Schultz2010}.  The implicit assumption that human judgment is monolithic and characterized by consensus should be carefully examined in future work.

The AI tools used in this study were all general-purpose tools.  Specialized tools, both open source and commercial, have been and are being developed to deal specifically with searching the literature and writing literature reviews \citep[e.g.,][]{Asai2026,ZabaletaLin2026}.  This study has not compared our results to these specialized tools.

%%%%%%%%%%%%%%%%%%%%%%%%%%%%%%%%%%%%%%%%%%%%%%%%%%%%%%%%%%%%%%%%%%%%%
\section{Conclusions}\label{sec:conclusions}

For better or worse, LLMs are now used to help write literature reviews. This study tested how well they do it, by following the process of writing a review step by step against a published review of the Spanish plume whose 102 articles are known. Five AI tools searched for the articles. Three were then given all of the 102 articles and asked to build a database, to write a synthesis, and to propose future work. Two human experts scored the syntheses without knowing which tool had written which.

The first question was whether the tools could find the relevant articles. The answer depended mainly on access. Without access to the university subscriptions, no tool found more than half of the articles; with access, the two coding agents found far more. But, a subscription was not the only obstacle. Some websites refused the tools' automated requests even for articles that were free to read, and one database searched only titles, abstracts and keywords, so it missed articles that mentioned the term ``Spanish plume" only in the body of the article. The tools also found recent articles more often than older ones.

The second question was whether the tools could interpret each article. The tools extracted text accurately; nearly all the quotations in their databases were word for word. They were less reliable at judging what an article did with the term ``Spanish plume". All three recognized the 18 articles that discussed the Spanish plume substantially, but they also classed many of the passing mentions as substantial. The tools often mistook a mention of the term for a discussion of it.

The third question was whether the tools could synthesize across articles. The syntheses were mostly accurate, but both reviewers found little insight in them. The tools overstated how far the literature agreed, and one repeated a description of the plume that appeared in several articles but conflicted with the original definition. Only an expert would have caught that. The tool that cited the most articles cited all the 18 substantial ones, but only by citing nearly everything. Thus, citing more did not necessarily mean choosing sources better. This result was likely constrained by the prompt, which was written to prevent errors and did not ask for interpretation.

The fourth question was whether the tools could propose future work. The tools did better here than might be expected. Two tools proposed research gaps close to those in the benchmark review, whereas the third repeated gaps the articles had already stated. Their weakness was focus, as all three tools proposed gaps on subjects broader than just the Spanish plume.

Taken together, these results show that the tools could help with a literature review but cannot be left to write one alone, at least at this time, results consistent with previous studies \citep[e.g.,][]{Mostafapour2024, Schiller2024}. They can find many of the relevant articles, extract what the articles say, and summarize it reasonably accurately, which potentially could save time. But their searches have gaps, their judgments of importance are unreliable, and they exhibit equally and often unwarranted confidence whether or not they are accurately reflecting the real state of the literature. Section~\ref{sec:recs} gives recommendations for using AI tools for constructing literature reviews, which come down to giving the tool the access it needs and checking its work at every step.

There are indications that different subjects may yield different results \citep{Tang2025}. Our conclusions in this study come from one topic, one run of each tool and two reviewers.  The AI tools are changing quickly. Given the biases in finding literature in our study and others, access in the future may influence citation rates as more authors use and cite references provided by AI tools.  The method can be repeated, however. Any published review with a known set of articles can serve as a benchmark, and repeating this test on other topics and with later versions of the tools would show how far these results hold.

%Given our mixed results in the value of the LLMs, what is our take-away message? Will an automatically generated literature review by an LLM achieve the tasks that we expect PhD students to learn?  Does the LLM bring insights that an experienced researcher reading and synthesizing the literature would achieve?  It may help in some respects, but on the whole the process still requires human input.  We hear that LLMs will continue to improve in time, and no doubt they will.  But, the process of researching literature, reading it, reflecting on it, and writing up a synthesis of it as the basis for motivating a research project remains essential to situating and motivating a research project. Outsourcing parts of the process of writing a literature review does a huge disservice to PhD students who are expected to learn the context of their research.  If they fail to understand the literature beyond a cursory reading of it as witnessed by the syntheses of the Spanish plume literature produced by these three LLMs, then students and researchers will be unable to address the context for their research within the breadth of the field, to any depth.

Given our mixed results on the value of the AI tools, what is our take-away message? The tools can help with parts of a literature review, but they did not bring the insight that an experienced researcher brings to reading and synthesizing a literature. LLMs will no doubt continue to improve. But the process of researching the literature, reading it, reflecting on it, and writing up a synthesis of it remains essential to situating and motivating a research project. Outsourcing parts of the process of writing a literature review does a huge disservice to PhD students and researchers, who are expected to learn the context of their research. If they understand the literature no more deeply than the syntheses of the Spanish plume literature produced by these three tools, they will be unable to place their research within the breadth of the field, to any depth.

%\clearpage
\acknowledgments
The authors acknowledge financial support from the N8 Centre of Excellence in Computationally Intensive Research (N8 CIR).
%as an N8 CIR intern and was financed by N8 CIR as part of the N8 Research Partnership. 
This work was carried out as a summer research internship at the University of Manchester and made use of the University's Computational Shared Facility and the N8 computing facilities. Author CRediT statement: AAK: methodology, software, validation, investigation, data curation, writing -- original draft, and visualization; DMS and AKAF: conceptualization, methodology, validation, writing -- review \& editing, supervision, and funding acquisition; ZZ: conceptualization, methodology, writing -- review \& editing, supervision, and funding acquisition; CM: conceptualization, methodology, writing -- review \& editing, and supervision.

\datastatement
The 102-article corpus can be found at \citet{Schultz2024}. The public archive of the present research is available from Zenodo (\url{https://doi.org/10.5281/zenodo.23849039}) \citep{AlKalbani2026} and GitHub (\url{https://github.com/abx40/spanish-plume-llm-literature-review}). It is organized by experiment and contains the prompts, raw model outputs, anonymized per-theme reviewer ratings, cleaned per-paper outcomes and summary tables, the data and images for the figures that show results, and scripts that redraw the figures and recompute the reported numbers. The archive is released under a CC BY 4.0 licence. Publisher-hosted source-article PDFs, the benchmark's full database, private authentication records, and named reviewer correspondence are not posted. The source PDFs and other non-public materials are held by the corresponding author; access is subject to source and reviewer permissions.

\bibliographystyle{ametsocV6}
\bibliography{references}

\end{document}